\documentclass[10pt,twocolumn,letterpaper]{article}

\usepackage[pagenumbers]{cvpr} 

\usepackage{graphicx}
\usepackage{amsmath}
\usepackage{amssymb}
\usepackage{booktabs}

\usepackage{graphicx}
\usepackage{amsmath}
\usepackage{amssymb}
\usepackage{siunitx}
\usepackage{ctable}
\usepackage{hhline}
\usepackage{subcaption}
\usepackage{csquotes}
\usepackage{multirow}
\usepackage{pifont}
\usepackage[super]{nth}
\usepackage{fixltx2e}
\usepackage{wrapfig}
\usepackage{enumitem}

\usepackage{xcolor}

\definecolor{darkergreen}{RGB}{21, 152, 56}
\definecolor{red2}{RGB}{252, 54, 65}
\newcommand{\cmark}{\textcolor{darkergreen}{\ding{51}}}%
\newcommand{\xmark}{\textcolor{red2}{\ding{55}}}%

\newcolumntype{L}[1]{>{\raggedright\let\newline\\\arraybackslash\hspace{0pt}}m{#1}}
\newcolumntype{C}[1]{>{\centering\let\newline\\\arraybackslash\hspace{0pt}}m{#1}}
\newcolumntype{R}[1]{>{\raggedleft\let\newline\\\arraybackslash\hspace{0pt}}m{#1}}

\usepackage[pagebackref,breaklinks,colorlinks]{hyperref}

\usepackage[capitalize]{cleveref}
\crefname{section}{Sec.}{Secs.}
\Crefname{section}{Section}{Sections}
\Crefname{table}{Table}{Tables}
\crefname{table}{Tab.}{Tabs.}

\def\confName{CVPR}
\def\confYear{2023}

\begin{document}

\title{Three Recipes for Better 3D Pseudo-GTs of \\3D Human Mesh Estimation in the Wild}

\author{
Gyeongsik Moon$^1$ \hspace{0.5cm} Hongsuk Choi$^2$ \hspace{0.5cm} Sanghyuk Chun$^3$ \hspace{0.5cm} Jiyoung Lee$^3$ \hspace{0.5cm} Sangdoo Yun$^3$\\
\\
$^1$ Meta Reality Labs \hspace{0.5cm} $^2$ Samsung AI Center - New York \hspace{0.5cm} $^3$ NAVER AI Lab\\
{\tt\small mks0601@gmail.com} \hspace{0.5cm} {\tt\small redstonepo@gmail.com} \hspace{0.5cm} {\tt\small \{sanghyuk.c, lee.j, sangdoo.yun\}@navercorp.com}
}
\maketitle

\begin{abstract}
Recovering 3D human mesh in the wild is greatly challenging as in-the-wild (ITW) datasets provide only 2D pose ground truths (GTs).
Recently, 3D pseudo-GTs have been widely used to train 3D human mesh estimation networks as the 3D pseudo-GTs enable 3D mesh supervision when training the networks on ITW datasets.
However, despite the great potential of the 3D pseudo-GTs, there has been no extensive analysis that investigates which factors are important to make more beneficial 3D pseudo-GTs.
In this paper, we provide three recipes to obtain highly beneficial 3D pseudo-GTs of ITW datasets.
The main challenge is that only 2D-based weak supervision is allowed when obtaining the 3D pseudo-GTs.
Each of our three recipes addresses the challenge in each aspect: depth ambiguity, sub-optimality of weak supervision, and implausible articulation.
Experimental results show that simply re-training state-of-the-art networks with our new 3D pseudo-GTs elevates their performance to the next level without bells and whistles.
The 3D pseudo-GT is publicly available\footnote{\url{https://github.com/mks0601/NeuralAnnot_RELEASE}}.
\end{abstract}

\section{Introduction}

3D human mesh estimation aims to localize 3D human mesh vertices in the 3D space.
The major challenge is the lack of 3D ground truths (GTs) of in-the-wild (ITW) datasets~\cite{johnson2011learning,lin2014microsoft,andriluka20142d}.
Images of ITW datasets are captured with a single camera without special equipment, such as inertial measurement units (IMUs) and multiple calibrated cameras, as ITW images are taken in our daily life.
As such special equipment is necessary to obtain 3D mesh data, only sparse 2D GT poses (\textit{i.e.}, 2D GT coordinates of about twenty joints) are available in ITW datasets without 3D dense mesh GTs that have thousands of vertices.

\begin{figure}[t]
\begin{center}
\includegraphics[width=\linewidth]{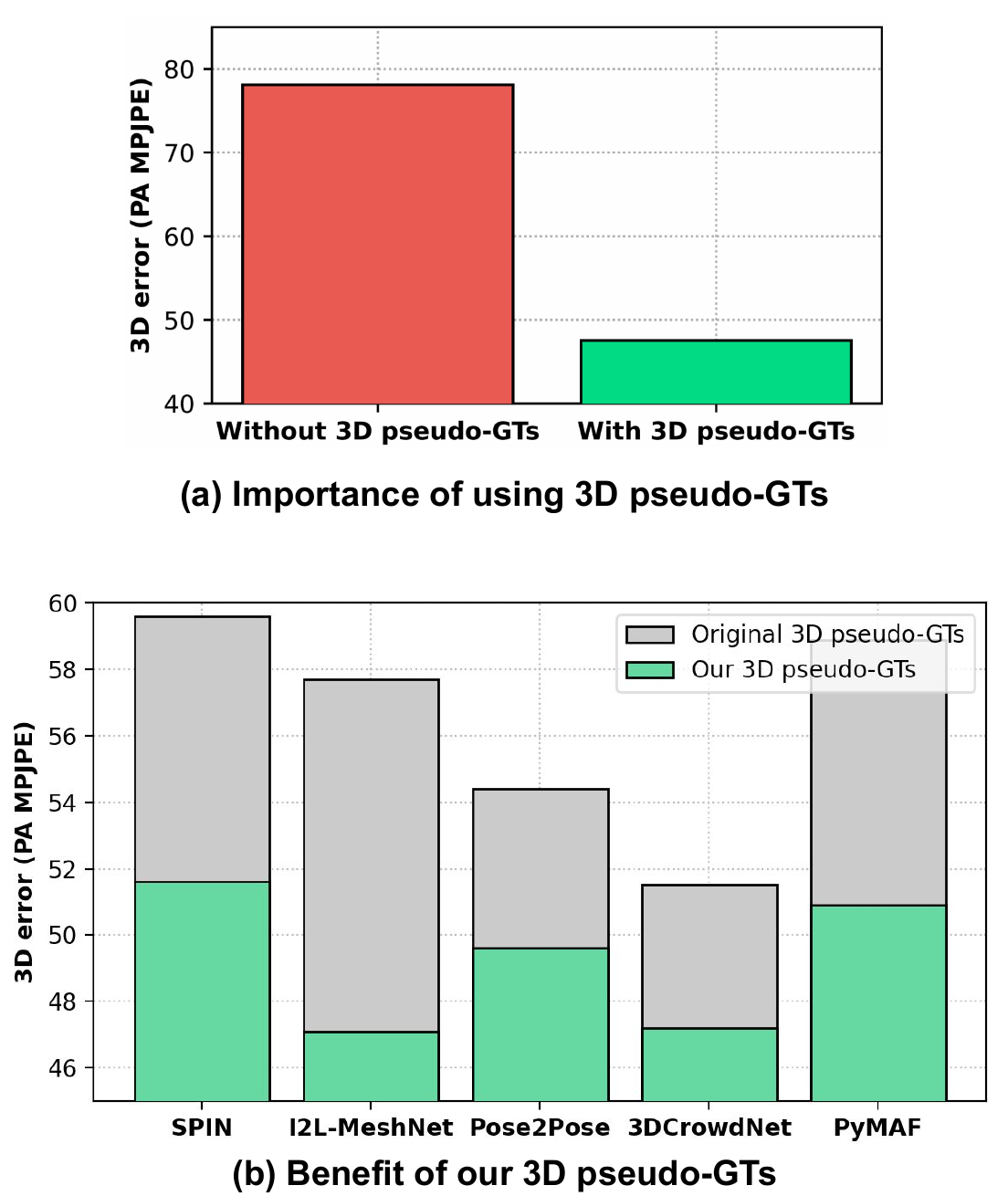}
\end{center}
\vspace*{-5mm}
\caption{
(a) 3D error (PA MPJPE) comparison on 3DPW~\cite{von2018recovering} between Pose2Pose~\cite{moon2022hand4whole} trained without and with 3D pseudo-GTs.
(b) 3D error (PA MPJPE) comparison on 3DPW~\cite{von2018recovering} between networks trained with their and our 3D pseudo-GTs.
The numbers are from Table~\ref{table:compare_annotation_various}.
}
\vspace*{-3mm}
\label{fig:intro}
\end{figure}

\begin{figure*}[t]
\begin{center}
\includegraphics[width=\linewidth]{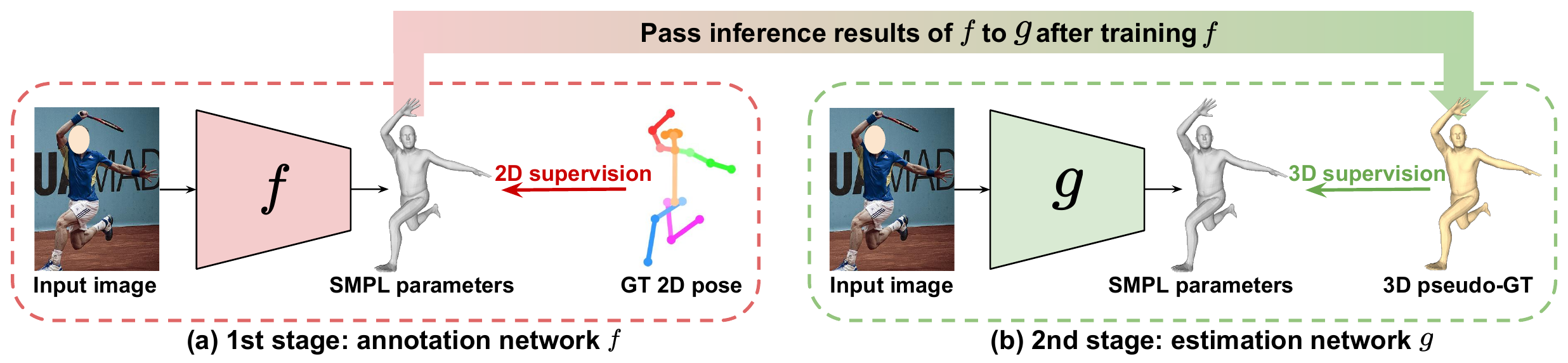}
\end{center}
\vspace*{-5mm}
\caption{
The overall pipeline of the proposed framework.
(a): In the first stage, the annotation network $f$ outputs SMPL parameters, \emph{weakly supervised with 2D GT pose}.
(b): After finishing training the annotation network $f$, the inference results of $f$ on seen training sets become 3D pseudo-GTs.
In the second stage, the estimation network $g$ is \emph{fully supervised with the 3D pseudo-GTs}.
For simplicity, we do not depict 3D supervisions of MoCap datasets during the mixed-batch training of $f$ and $g$.
}
\label{fig:overall_pipeline}
\end{figure*}

The main training strategy for the 3D human mesh estimation in the wild is a mixed-batch training~\cite{kocabas2021pare,ROMP,moon2022hand4whole,choi2022learning,moon2020i2l,choi2020p2m,choi2021beyond,lin2021end,lin2021mesh,kocabas2020vibe}, which takes half samples of a mini-batch from motion capture (MoCap) datasets~\cite{ionescu2014human3,mehta2017monocular,joo2015panoptic,moon2020interhand2,yu2020humbi} and rest samples from ITW datasets.
MoCap datasets are captured from a controlled environment, such as a lab or studio, and they provide 3D pose and mesh GTs by utilizing special equipment, such as multiple calibrated cameras.
During the mixed-batch training, samples from MoCap datasets are supervised with 3D GT meshes, and those from ITW datasets are supervised with 3D pseudo-GT meshes.
The contribution of MoCap datasets is providing 3D supervision with their accurate 3D GTs, which do not exist in ITW datasets.
However, using MoCap datasets is not sufficient for the best performance in the wild.
This is because they are collected from the controlled environment; therefore, their image appearances, such as illumination and backgrounds, are highly limited and far from those of ITW datasets~\cite{moon20223d,choi2020p2m,choi2022learning}.
To cope with such limitation, 3D pseudo-GTs of ITW datasets have been widely used to provide 3D supervision to ITW samples.
Although 3D pseudo-GTs contain errors in nature, they provide 3D supervision to ITW samples,  which can complement 2D-based weak supervision from 2D GT poses of ITW datasets.

Fig.~\ref{fig:intro} (a) shows that the 3D pseudo-GTs of ITW datasets boost the performance a lot compared to a counterpart that does not utilize the 3D pseudo-GTs.
The figure shows that 3D pseudo-GTs are greatly important for high performance and justifies the \textbf{two-stage training pipeline} (Fig.~\ref{fig:overall_pipeline}), of which 
the \textbf{first stage} is acquiring 3D pseudo-GTs, and the \textbf{second stage} is training a 3D human mesh estimation network~\cite{kocabas2021pare,ROMP,moon2022hand4whole,choi2022learning,moon2020i2l,choi2020p2m,choi2021beyond,lin2021end,lin2021mesh,kocabas2020vibe} with the 3D pseudo-GTs.
In the first stage, the 3D pseudo-GTs are acquired using either the iterative fitting framework~\cite{bogo2016keep,pavlakos2019expressive} or \emph{external} annotation network~\cite{kolotouros2019learning,joo2021eft,moon2022neuralannot}.
We denote the annotation network of the first stage by $f$ and the estimation network of the second stage by $g$.

Annotation networks $f$~\cite{bogo2016keep,pavlakos2019expressive,kolotouros2019learning,joo2021eft,moon2022neuralannot} are weakly supervised with 2D GT poses to obtain 3D pseudo-GTs of ITW datasets.
The weak supervision of ITW samples is enabled by SMPL body model~\cite{loper2015smpl}, which produces 3D human mesh from pose and shape parameters in a differentiable way.
After extracting 3D joint coordinates from the 3D mesh and projecting them to the 2D space, the 2D-based weak supervision minimizes the distance between the projected 2D joint coordinates and 2D GT pose.
In this way, the 2D GT pose weakly supervises SMPL parameters, which can make all vertices of the 3D mesh fit to the 2D GT pose.
In this paper, we define 3D pseudo-GTs as SMPL parameters.

Unfortunately, although many recent 3D human mesh estimation methods train their networks $g$~\cite{kocabas2021pare,ROMP,moon2022hand4whole,choi2022learning,moon2020i2l,choi2020p2m,choi2021beyond,lin2021end,lin2021mesh,kocabas2020vibe} with 3D pseudo-GTs of ITW datasets for their performances, there has been no extensive analysis that investigates which factors are important to obtain beneficial 3D pseudo-GTs.
\textbf{In this paper, we provide three recipes for highly beneficial 3D pseudo-GTs of ITW datasets.}
The main challenge is that \emph{only 2D-based weak supervision is allowed} without 3D evidence in ITW datasets when obtaining the 3D pseudo-GTs.
The absence of the 3D evidence when training the annotation networks $f$ (\textit{i.e.}, the first stage in Fig.~\ref{fig:overall_pipeline}) causes severe ambiguities, while the estimation networks $g$ (\textit{i.e.}, the second stage in Fig.~\ref{fig:overall_pipeline}) suffer less from them as the 3D pseudo-GTs from the first stage serve 3D evidence.

We address the challenge of obtaining beneficial 3D pseudo-GTs (\textit{i.e.}, the first stage in Fig.~\ref{fig:overall_pipeline}) in three aspects: \emph{depth ambiguity, sub-optimality of weak supervision, and implausible articulation.}
First, multiple 3D data (\textit{e.g.}, SMPL parameters) corresponds to the same 2D evidence, which incurs depth ambiguity. 
Second, weak supervision signals make networks converge to sub-optimal points compared to full supervision. 
Finally, 3D human meshes with anatomically implausible articulations can correspond to the 2D GT pose.
All the previous iterative fitting frameworks~\cite{bogo2016keep,pavlakos2019expressive} and annotation networks $f$~\cite{kolotouros2019learning,joo2021eft,moon2022neuralannot} suffer from the problems as they rely on the 2D-based weak supervision when obtaining 3D pseudo-GTs; however, they have not carefully considered the problems.
Fig.~\ref{fig:intro} (b) shows that without bells and whistles, simply re-training state-of-the-art estimation networks $g$ with our new 3D pseudo-GTs elevate their performance to the next level on ITW benchmarks~\cite{von2018recovering}.
Fig.~\ref{fig:recipe} shows that the performance of estimation network $g$ improves with each recipe applied.
We will publicly open our 3D pseudo-GTs, which can benefit the community and following works.

\begin{table*}[t]
\footnotesize
\centering
\setlength\tabcolsep{1.0pt}
\def\arraystretch{1.1}
\begin{tabular}{C{3.5cm}|C{3.0cm}|C{3.8cm}|C{4.0cm}}
\specialrule{.1em}{.05em}{.05em}
Annotation networks $f$ & Train $f$ on 3DPW & Initialization of $f$ & Use VPoser and L2 reg.  in $f$ \\ \hline
SPIN~\cite{kolotouros2019learning} & \xmark & ImageNet classification~\cite{he2016deep} &  \xmark\\
EFT~\cite{joo2021eft} & \xmark  & 3D pose network~\cite{kolotouros2019learning} &\xmark  \\
NeuralAnnot~\cite{moon2022neuralannot} & \xmark & ImageNet classification~\cite{he2016deep} & \cmark \\
Ours & \cmark & 2D pose network~\cite{xiao2018simple} & \cmark \\ \specialrule{.1em}{.05em}{.05em}
\end{tabular}
\vspace*{-2mm}
\caption{
Comparison of previous annotation networks and ours.
}
\label{table:compare_annotation_network_novelty}
\vspace{-2em}
\end{table*}

\section{3D pseudo-GTs of ITW datasets}

\subsection{Overall pipeline}
Fig.~\ref{fig:overall_pipeline} shows the overall pipeline of the proposed framework.
Our entire system consists of two networks: annotation network $f$ and estimation network $g$, where both networks are trained with the mixed-batch training strategy.
The annotation network $f$ is trained with 2D and 3D GTs of ITW and MoCap datasets, respectively.
Please note that the mixed-batch training of the annotation network $f$ is different from that of estimation network $g$ in that only 2D supervision, without 3D supervision, is available for ITW samples.
The testing results of $f$ on seen training images of ITW datasets become 3D pseudo-GTs.
The 3D pseudo-GTs are used to train the estimation network $g$.
As developing a new network architecture is not our focus, we design the annotation network $f$ to have the network architecture of Pose2Pose~\cite{moon2022hand4whole}, a state-of-the-art SMPL parameter regression network.
For the details of Pose2Pose, please refer to the supplementary material.
We use various state-of-the-art 3D human mesh estimation networks~\cite{moon2020i2l,moon2022hand4whole,choi2022learning,kocabas2021pare,kolotouros2019learning,lin2021end} for the estimation network $g$ and show generalizability of our 3D pseudo-GTs to them in the experimental section.

\begin{figure}[t]
\begin{center}
\includegraphics[width=\linewidth]{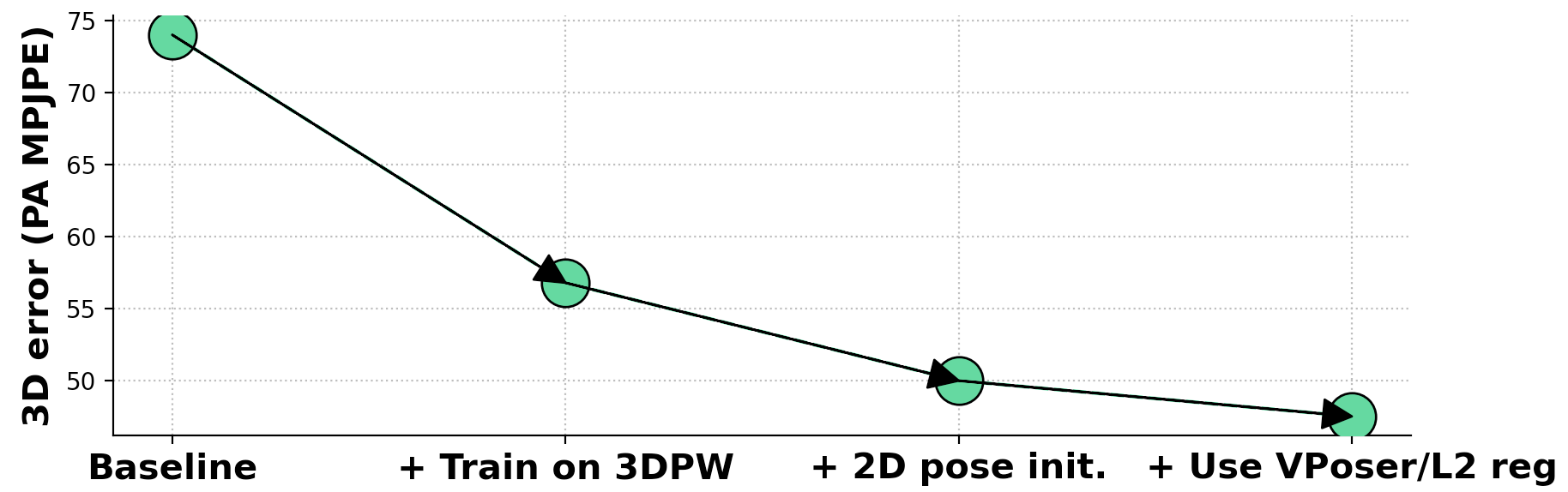}
\end{center}
\vspace*{-5mm}
\caption{
3D errors (PA MPJPE) of estimation network $g$, trained with different 3D pseudo-GTs, on 3DPW~\cite{von2018recovering}. 
The lower the better.
Each pseudo-GT is designed for mitigating depth ambiguity, sub-optimality of weak supervision, and implausible articulation. 
Our three recipes significantly improve 3D errors of $g$.
}
\vspace*{-3mm}
\label{fig:recipe}
\end{figure}

\subsection{Three recipes for 3D pseudo-GTs}
The major challenge to obtaining beneficial 3D pseudo-GTs of ITW datasets is that only weak supervision targets (\textit{i.e.}, 2D GT poses) are available without 3D evidence.
The absence of the 3D evidence when training the annotation networks $f$ (\textit{i.e.}, the first stage in Fig.~\ref{fig:overall_pipeline}) causes severe ambiguities, while the estimation networks $g$ (\textit{i.e.}, the second stage in Fig.~\ref{fig:overall_pipeline}) suffers less from the ambiguities as the 3D pseudo-GTs from the first stage serve 3D evidence.
We design our recipes to address the challenge of obtaining more beneficial 3D pseudo-GTs in three aspects: \emph{depth ambiguity, sub-optimality of weak supervision, and implausible articulation}.
Fig.~\ref{fig:recipe} shows how the 3D error of the estimation network $g$ changes when the 3D pseudo-GTs of ITW datasets are obtained following our recipes.
Our three recipes are summarized below.

\noindent\textbf{\textit{1. To resolve the depth ambiguity,  even if the scales of datasets are small, collect ITW datasets with 3D GTs (\textit{e.g.}, 3DPW~\cite{von2018recovering}) and train the annotation network \boldmath{$f$} on them.}}
The 2D-based weak supervision causes depth ambiguity as there can be an infinite number of 3D data (\textit{e.g.}, SMPL parameters) that correspond to the same 2D evidence.
Previous annotation networks $f$~\cite{kolotouros2019learning,joo2021eft} alleviated the depth ambiguity by using MoCap datasets during the mixed-batch training.
As MoCap datasets provide 3D GT meshes, their networks learn an image-to-3D mesh function from MoCap datasets, and the learned function is shared with the ITW case in the same network.
However, it is not sufficient as MoCap images have largely different image appearances, such as backgrounds, illuminations, and colors, compared to those of ITW images.
The reason for such a large appearance gap is that MoCap datasets are captured from a restricted environment, such as a studio or lab, while ITW datasets are captured from anywhere in our daily life.
Due to such a large appearance gap, knowledge learned from MoCap samples might not sufficiently be transferred to the ITW case.

\begin{figure}[t]
\begin{center}
\includegraphics[width=\linewidth]{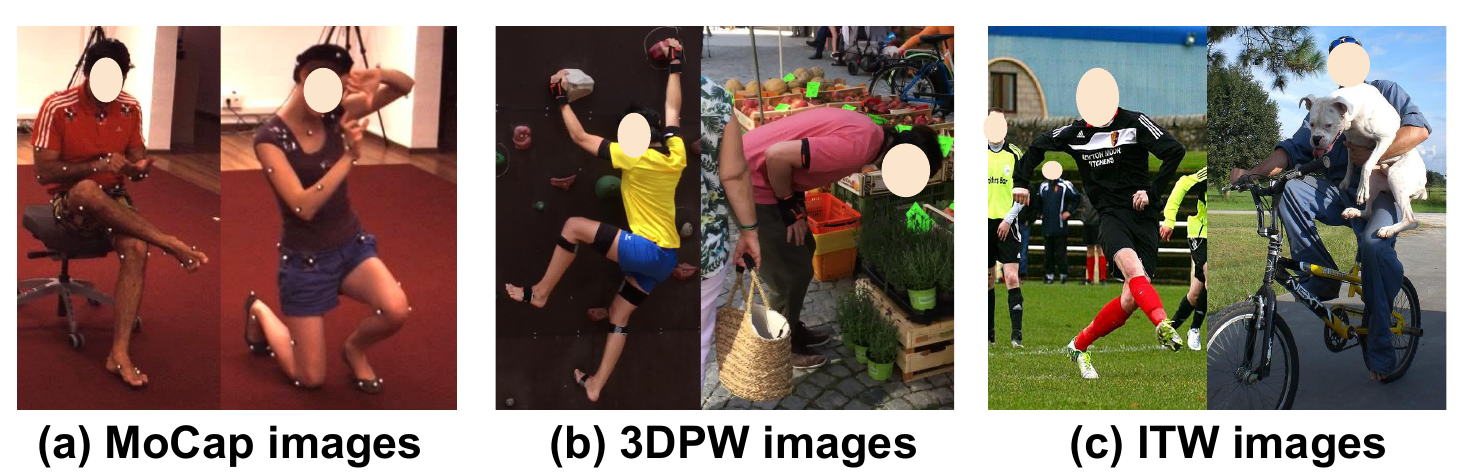}
\end{center}
\vspace*{-5mm}
\caption{
Comparisons of images from MoCap dataset~\cite{ionescu2014human3}, 3DPW~\cite{von2018recovering}, and ITW dataset~\cite{lin2014microsoft}.
}
\vspace*{-3mm}
\label{fig:3dpw_compare}
\end{figure}

To bridge MoCap and ITW datasets, even if the scales of datasets are small, we propose to collect ITW datasets with 3D GTs and train the annotation network $f$ on them.
One example of such a small-scale ITW dataset with 3D GTs is 3DPW~\cite{von2018recovering}.
3DPW is captured from the outdoor environment with moving cameras, and its image appearance is much closer to those of ITW images than existing MoCap datasets~\cite{ionescu2014human3,mehta2017monocular,joo2015panoptic,moon2020interhand2,yu2020humbi}, as shown in Fig.~\ref{fig:3dpw_compare}.
Importantly, it provides accurate 3D GTs thanks to IMUs, attached to subjects' bodies and hidden under clothes.
Therefore, \emph{the 3DPW dataset serves as a bridge between MoCap and ITW datasets}.
None of previous annotation networks $f$~\cite{kolotouros2019learning,joo2021eft,moon2022neuralannot} is trained on such small-scale ITW datasets with 3D GTs; instead, some of them~\cite{kolotouros2019learning,joo2021eft} are trained on additional ITW datasets with 2D GTs~\cite{andriluka20142d,johnson2011learning}.
We observed that despite its small scale (23K unique images), utilizing 3DPW as an additional training set to train the annotation network $f$ improves the 3D pseudo-GTs of ITW datasets a lot, which results in lower 3D errors of the estimation networks $g$ on multiple 3D benchmarks~\cite{von2018recovering,mehta2018single}.
On the other hand, we show that \emph{95$\times$ larger ITW datasets (2.2M unique images~\cite{kanazawa2019learning})} with 2D GTs are not helpful for the 3D pseudo-GTs.
This implies that the existence of 3D GTs in 3DPW is much more important to make 3D pseudo-GTs better than a large number of 2D GTs and rich appearance distribution from ITW datasets.
Please note that the advantage of 3DPW for the annotation network $f$ is not from the in-domain similarity between the 3DPW training and testing set.
Although we use the 3DPW training set when training annotation networks $f$ to obtain 3D pseudo-GTs, the performance of the estimation network $g$ improves on multiple benchmarks without using 3DPW for the training of $g$.

\noindent\textbf{\textit{2. To resolve the sub-optimality of weak supervision, initialize the annotation network \boldmath{$f$} with a pre-trained 2D pose estimation network.}}
When training the annotation network $f$, samples from ITW datasets are weakly supervised with 2D GTs without 3D supervision.
The weak supervision might make networks converge to sub-optimal points as it involves ambiguity in nature compared to the full supervision~\cite{bilen2016weakly,durand2017wildcat,tang2018pcl,wan2019c,choe2020wsoleval,choe2022evaluation}.
We alleviate the sub-optimality by initializing ResNet backbone~\cite{he2016deep} of our annotation network $f$ with that of a pre-trained 2D pose estimation network~\cite{xiao2018simple}.
From the perspective of the representation learning~\cite{he2020momentum,grill2020bootstrap,henaff2021efficient}, the pre-trained 2D pose estimation network can extract human articulation information much better than the random initialization and ImageNet~\cite{russakovsky2015imagenet} classification network~\cite{he2016deep}.
By extracting useful human articulation features from images at the early stage of the training, our annotation network $f$ can reach a better convergence point, which results in more beneficial 3D pseudo-GTs.

\noindent\textbf{\textit{3. To resolve the implausible articulation, use a combination of VPoser~\cite{pavlakos2019expressive} and L2 regularizer in the annotation network \boldmath{$f$}.}}
When training the annotation network $f$, samples from ITW datasets are supervised only with 2D GTs without 3D targets (\textit{i.e.}, 3D GTs and 3D pseudo-GTs).
However, relying only on the 2D-based data term might make the networks produce 3D meshes with anatomically implausible articulations (\textit{e.g.}, penetration and out of possible range of 3D joint rotations) as such 3D meshes can also minimize the 2D-based data term.
To prevent this, we use a combination of VPoser~\cite{pavlakos2019expressive} and L2 regularizer when training the annotation network $f$.
VPoser is a variational auto-encoder, which embeds large-scale SMPL pose parameters~\cite{mahmood2019amass} to a Gaussian latent space.
It can effectively limit 3D human meshes, produced from SMPL parameters, to anatomically plausible ones.
We modify our annotation network $f$ to estimate the latent code of VPoser as the original Pose2Pose network directly estimates SMPL pose parameter.
In addition, during the training, we newly apply an L2 regularizer to the estimated latent code to enforce the code to be in the latent space of VPoser.

\noindent\textbf{\textit{* Novelty of our recipes.}}
Although all three recipes can be applied to the estimation network $g$, we observed that the effect of our recipes is much larger when they are applied to annotation networks $f$ compared to being applied to estimation networks $g$.
This is because annotation networks $f$ do not have 3D evidence of ITW datasets in the training stage, while estimation networks $g$ utilize 3D pseudo-GTs as 3D evidence of ITW datasets.
Therefore, annotation networks $f$ suffer from the three ambiguities, while estimation networks $g$ suffer much less.

Table~\ref{table:compare_annotation_network_novelty} shows a comparison of previous annotation networks and ours.
Although NeuralAnnot~\cite{moon2022neuralannot} used VPoser~\cite{pavlakos2019expressive} like ours, they did not investigate that the usage of VPoser is especially helpful for the annotation network $f$, while has a small effect when VPoser is used for the estimation network $g$.
We show this analysis in the experimental section, which indicates that the usage of VPoser is specially designed for the annotation network $f$.
\section{Experiment}

\subsection{Datasets}
\noindent\textbf{MoCap datsets.}
We use Human3.6M (H36M)~\cite{ionescu2014human3} and MPI-INF-3DHP (MI)~\cite{mehta2017monocular} as MoCap datasets.
They are used only to train both the annotation network $f$ and estimation network $g$ and are not used for evaluation purposes as our goal is an evaluation on ITW benchmarks, not on MoCap ones.

\noindent\textbf{ITW datasets with 2D GTs.}
We use COCO~\cite{lin2014microsoft}, MPII~\cite{andriluka20142d}, and LSPET~\cite{johnson2011learning} as ITW datasets, which provide 2D GTs.
They are used for the training of annotation networks $f$ and estimation networks $g$.
The inference results of annotation networks $f$ on the above ITW datasets become 3D pseudo-GTs, used to train estimation networks $g$.
The above ITW datasets are not used for evaluation purposes as they do not provide 3D targets.

\noindent\textbf{ITW datasets with 3D GTs.}
We use 3DPW~\cite{von2018recovering} and MuPoTS~\cite{mehta2018single} as additional ITW datasets.
Both contain images, captured from outdoor, with 3D GTs thanks to IMUs or multi-view marker-less motion capture systems.
3DPW training split is used to train annotation networks $f$ and optionally estimation networks $g$, and 3DPW test split used to evaluate $g$.
MuPoTS is used only for the evaluation purpose of estimation networks $g$.

\subsection{Evaluation protocol}
As the main focus of this paper is acquiring better 3D pseudo-GTs of ITW datasets, we evaluate how much 3D pseudo-GTs are beneficial for the estimation network $g$.
To this end, we first acquire 3D pseudo-GTs using an annotation network $f$.
Then, we train an estimation network $g$ using the mixed-batch training strategy, where 3D pseudo-GTs are from the annotation network $f$.
Finally, we report the most widely used 3D error metric in the 3D human mesh estimation community, PA MPJPE, of the estimation network $g$ on the multiple 3D ITW benchmark (\textit{i.e.}, test split of 3DPW and MuPoTS).
The errors are measured from 3D joint coordinates, extracted from 3D meshes following previous works~\cite{kolotouros2019learning,moon2022hand4whole}.
We additionally use 3DPCK as an evaluation metric of MuPoTS as previous works~\cite{jiang2020coherent,choi2022learning}.
The lower 3D errors or the higher 3DPCK of the estimation network $g$ indicate the better 3D pseudo-GTs from the annotation network $f$.

\begin{table}[t]
\footnotesize
\centering
\setlength\tabcolsep{1.0pt}
\def\arraystretch{1.1}
\begin{tabular}{C{4.2cm}|C{2.5cm}}
\specialrule{.1em}{.05em}{.05em}
Training sets of $g$ & 3D errors of $g$ \\ \hline
\multicolumn{1}{l|}{H36M+MI+[COCO]\textsubscript{SMPLify}} & 64.76 / 87.42 \\
\multicolumn{1}{l|}{H36M+MI+[COCO]\textsubscript{SMPLify-X}} & 60.40 / 81.64 \\
\multicolumn{1}{l|}{H36M+MI+[COCO]\textsubscript{SPIN}} & 60.70 / 80.24 \\
\multicolumn{1}{l|}{H36M+MI+[COCO]\textsubscript{EFT}} &  55.15 / 78.02 \\
\multicolumn{1}{l|}{H36M+MI+[COCO]\textsubscript{CLIFF}} &  53.36 / 75.59 \\
\multicolumn{1}{l|}{H36M+MI+[COCO]\textsubscript{NeuralAnnot}} & 53.34 / 76.98 \\ \hline
\multicolumn{1}{l|}{H36M+MI+[COCO]\textsubscript{Ours wo. first recipe}} & 50.82 / 75.63 \\
\multicolumn{1}{l|}{H36M+MI+[COCO]\textsubscript{Ours wo. second recipe}} & 48.84 / 75.72 \\
\multicolumn{1}{l|}{H36M+MI+[COCO]\textsubscript{Ours wo. third recipe}} & 48.31 / 75.70 \\
\multicolumn{1}{l|}{H36M+MI+[COCO]\textsubscript{\textbf{Ours}}} &  \textbf{47.52} / \textbf{74.55} \\ \specialrule{.1em}{.05em}{.05em}
\end{tabular}
\vspace*{-2mm}
\caption{
Comparison of Pose2Pose trained with different 3D pseudo-GTs of COCO. 
For all settings, Pose2Pose is used as the estimation network $g$.
The subscript at the square brackets denotes the annotation network $f$ to obtain the 3D pseudo-GTs.
The left and right 3D errors of $g$ (PA MPJPE) are calculated on 3DPW and MuPoTS, respectively.
}
\label{table:compare_annotation_pose2pose}
\vspace{-1em}
\end{table}

\subsection{Comparison with state-of-the-art methods}

\begin{figure}[t]
\begin{center}
\includegraphics[width=\linewidth]{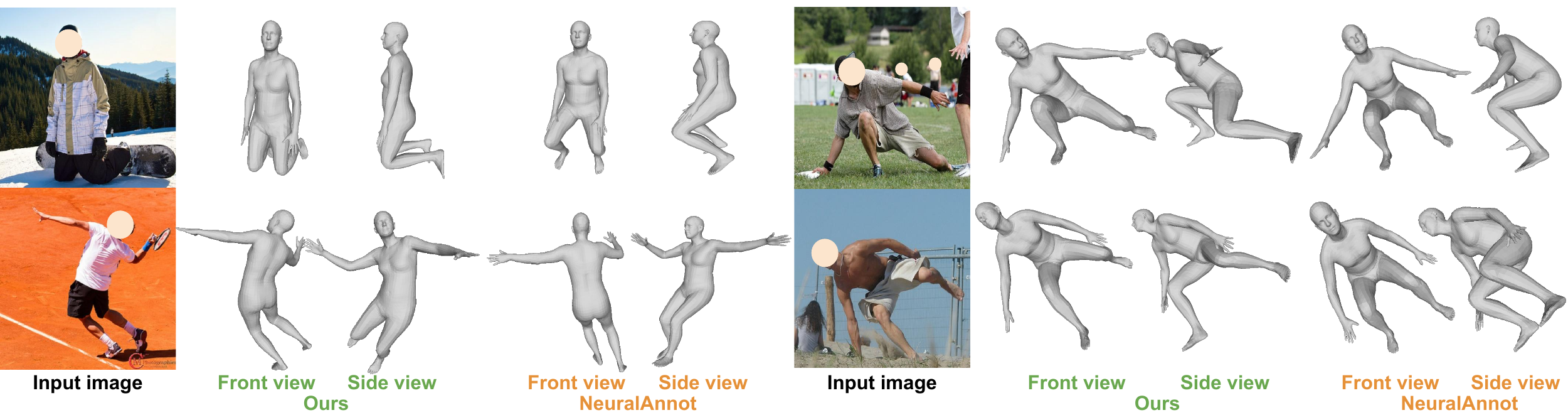}
\end{center}
\vspace*{-5mm}
\caption{
Visual comparison between 3D pseudo-GTs of COCO from ours and NeuralAnnot~\cite{moon2022neuralannot}.
}
\label{fig:qualitative_comparison}
\end{figure}

\noindent\textbf{Comparison with previous annotation networks \boldmath{$f$}.}
Table~\ref{table:compare_annotation_pose2pose} shows that Pose2Pose~\cite{moon2022hand4whole}, trained with 3D pseudo-GTs of COCO from our annotation network $f$, achieves the lowest 3D errors on both 3DPW and MuPoTS.
Even after we apply only two of three recipes, Pose2Pose trained with our $f$ 3D pseudo-GTs still outperforms counterparts trained with 3D pseudo-GTs from previous $f$.
For all settings, only 3D pseudo-GTs of COCO are different, and the remaining settings are the same.
This proves the superiority of our annotation network $f$ compared to previous annotation networks~\cite{kolotouros2019learning,joo2021eft,moon2022neuralannot,li2022cliff} and iterative fitting frameworks~\cite{bogo2016keep,pavlakos2019expressive} regarding the ability to acquire beneficial 3D pseudo-GTs.
For the comparison, we use the public 3D pseudo-GTs of previous works~\cite{kolotouros2019learning,joo2021eft,moon2022neuralannot,li2022cliff}.
The 3D pseudo-GTs of SMPLify~\cite{bogo2016keep} are provided in the websites of SPIN, and those of SMPLify-X~\cite{pavlakos2019expressive} are obtained by running their official codes to 2D GT poses of COCO.
Fig.~\ref{fig:qualitative_comparison} visually demonstrates that our 3D pseudo-GTs are better than those of NeuralAnnot~\cite{moon2022neuralannot}.
NeuralAnnot fails to capture difficult poses, such as bent poses of the top-row examples.
In addition, it suffers from depth ambiguity, as shown in bottom-row examples.
In the bottom-left example, the right shoulder and hip should be farther from the camera than the left ones.
Also, in the bottom-right example, the left leg should be behind the right leg.
On the other hand, our 3D pseudo-GTs successfully capture such difficult cases.

\begin{table*}[t]
\footnotesize
\centering
\setlength\tabcolsep{1.0pt}
\def\arraystretch{1.1}
\begin{tabular}{C{3.2cm}|C{7.5cm}|C{3.0cm}}
\specialrule{.1em}{.05em}{.05em}
Estimation networks $g$ & Training sets of $g$ & 3D errors of $g$ \\ \hline
\multirow{2}{*}{SPIN~\cite{kolotouros2019learning}} & \multicolumn{1}{l|}{H36M+MI+[COCO+MPII+LSPET]\textsubscript{SMPLify}} & 59.6 (21.6/24.2/40.8) \\ 
& \multicolumn{1}{l|}{H36M+MI+[COCO+MPII+LSPET]\textsubscript{\textbf{Ours}}} & \textbf{51.6 (19.0/20.7/35.4)} \\ \hline
\multirow{2}{*}{I2L-MeshNet~\cite{moon2020i2l}} & \multicolumn{1}{l|}{H36M+MuCo+[COCO]\textsubscript{SMPLify-X}}  & 57.7 (20.6/21.7/40.8) \\
& \multicolumn{1}{l|}{H36M+MuCo+[COCO]\textsubscript{\textbf{Ours}}}  & \textbf{47.1 (17.1/18.5/32.6)} \\ \hline
\multirow{2}{*}{Pose2Pose~\cite{moon2022hand4whole}} & \multicolumn{1}{l|}{H36M+[COCO+MPII]\textsubscript{NeuralAnnot}} & 54.4 (19.1/20.3/39.0) \\ 
& \multicolumn{1}{l|}{H36M+[COCO+MPII]\textsubscript{\textbf{Ours}}} & \textbf{49.6 (18.4/18.8/34.7)} \\ \hline
\multirow{2}{*}{3DCrowdNet~\cite{choi2022learning}} & \multicolumn{1}{l|}{H36M+MuCo+CrowdPose+[COCO+MPII]\textsubscript{NeuralAnnot}} &  51.5 (17.6/18.2/36.3) \\ 
& \multicolumn{1}{l|}{H36M+MuCo+CrowdPose+[COCO+MPII]\textsubscript{\textbf{Ours}}}  & \textbf{47.2 (16.8/17.7/33.5)} \\ \hline
\multirow{2}{*}{PARE~\cite{kocabas2021pare}} & \multicolumn{1}{l|}{[COCO]\textsubscript{EFT}}  & 57.3 (20.3/20.3/41.7) \\ 
& \multicolumn{1}{l|}{[COCO]\textsubscript{\textbf{Ours}}}  & \textbf{47.3 (17.5/18.2/32.9)} \\ \hline
\multirow{2}{*}{PyMAF~\cite{zhang2021pymaf}} & \multicolumn{1}{l|}{H36M+MI+[COCO+MPII+LSPET]\textsubscript{SPIN}}  & 58.9 (21.0/23.7/41.8) \\ 
& \multicolumn{1}{l|}{H36M+MI+[COCO+MPII+LSPET]\textsubscript{\textbf{Ours}}}  & \textbf{50.9 (18.0/20.8/35.1)} \\ \hline
\multirow{2}{*}{METRO~\cite{lin2021end}} &  \multicolumn{1}{l|}{H36M+UP3D+MuCo+3DPW+MPII+[COCO]\textsubscript{SMPLify-X}} & 47.9 (18.8/18.5/32.4) \\
& \multicolumn{1}{l|}{H36M+UP3D+MuCo+3DPW+MPII+[COCO]\textsubscript{\textbf{Ours}}} &  \textbf{45.8 (17.9/17.2/31.3)} \\ 
\specialrule{.1em}{.05em}{.05em}
\end{tabular}
\vspace*{-2mm}
\caption{
Comparison of various estimation networks $g$, trained with different 3D pseudo-GTs of ITW datasets. 
Notations are the same as Table~\ref{table:compare_annotation_pose2pose}, except the 3D errors of $g$ (PA MPJPE) are calculated on 3DPW, and three errors in the parenthesis are from $x$-, $y$-, and $z$-axis, respectively.
}
\label{table:compare_annotation_various}
\vspace{-1.5em}
\end{table*}

\begin{table}[t]
\footnotesize
\centering
\setlength\tabcolsep{1.0pt}
\def\arraystretch{1.1}
\begin{tabular}{C{3.5cm}|C{2.0cm}}
\specialrule{.1em}{.05em}{.05em}
Estimation networks $g$ & 3D errors of $g$ \\ \hline
SPIN~\cite{kolotouros2019learning}  & 59.2 \\
Pose2Mesh~\cite{choi2020p2m}  & 58.9 \\
PyMAF~\cite{zhang2021pymaf} & 58.9 \\
I2L-MeshNet~\cite{moon2020i2l}  & 57.7 \\
Pose2Pose~\cite{moon2022hand4whole}  & 54.4 \\
ROMP~\cite{ROMP} & 53.3 \\
3DCrowdNet~\cite{choi2022learning} & 51.5 \\
PARE~\cite{kocabas2021pare} & 49.3 \\
HybrIK~\cite{li2021hybrik}  & 48.8 \\
\textbf{Our 3DCrowdNet} & \textbf{46.1} \\ \hline
METRO*~\cite{lin2021end}  & 47.9 \\
PARE*~\cite{kocabas2021pare} & 46.4 \\
MeshGraphormer*~\cite{lin2021mesh} & 45.6 \\ 
\textbf{Our 3DCrowdNet*} & \textbf{43.6} \\ \specialrule{.1em}{.05em}{.05em}
\end{tabular}
\vspace*{-2mm}
\caption{
Comparisons of 3D human mesh estimation methods on 3DPW.
* denotes additional training on 3DPW.
}
\label{table:compare_estimation_network_3dpw}
\vspace{-1em}
\end{table}

Table~\ref{table:compare_annotation_various} shows the generalizable benefits of our 3D pseudo-GTs to various state-of-the-art estimation networks $g$.
For the experiment, we train two networks for each estimation network $g$ using official codes of it: one with 3D pseudo-GTs of ITW datasets that it originally used, and the other with 3D pseudo-GTs of ITW datasets that are obtained from our annotation network $f$. 
Please note that other than 3D pseudo-GTs of ITW datasets, all other settings, such as the training schedule, remain the same for each estimation network $g$.
The table shows that simply changing 3D pseudo-GTs of ITW datasets from theirs to ours greatly decreases the 3D errors.
In particular, the error of the $z$-axis decreases the most among $x$-, $y$- and $z$-axis errors, which shows that our 3D pseudo-GTs effectively alleviate the depth ambiguity of the 3D human mesh estimation from a monocular image.
The reason for the relatively small $z$-axis error gap of METRO~\cite{lin2021end} is that it is additionally trained on 3DPW.
Nevertheless, our 3D pseudo-GTs still enhance its performance.
As detailed training set configurations of PARE~\cite{kocabas2021pare} are not publicly available, we simply trained the PARE network only on COCO, the reason for different 3D errors from their paper.

\begin{table}[t]
\footnotesize
\centering
\setlength\tabcolsep{1.0pt}
\def\arraystretch{1.1}
\begin{tabular}{C{3.5cm}|C{1.0cm}C{1.5cm}}
\specialrule{.1em}{.05em}{.05em}
\multirow{2}{*}{Estimation networks $g$} & \multicolumn{2}{c}{3DPCK of $g$} \\ 
& All & Matched \\ \hline
SMPLify-X~\cite{pavlakos2019expressive} & 62.8 & 68.0 \\
HMR~\cite{kanazawa2018end} & 66.0 & 70.9 \\
Jiang~et al.~\cite{jiang2020coherent}  & 69.1 & 72.2 \\
3DCrowdNet~\cite{choi2022learning} & 72.7 & 73.3 \\ 
\textbf{Our 3DCrowdNet} & \textbf{76.2} & \textbf{76.9} \\ 
\specialrule{.1em}{.05em}{.05em}
\end{tabular}
\vspace*{-2mm}
\caption{
Comparisons of 3D human mesh estimation methods on MuPoTS.
The higher the better.
}
\label{table:compare_estimation_network_mupots}
\vspace{-1em}
\end{table}

\noindent\textbf{Pushing the performance of state-of-the-art networks.}
Using our 3D pseudo-GTs of ITW datasets, we investigate how far state-of-the-art networks can become better.
To this end, we re-trained 3DCrowdNet~\cite{choi2022learning} with our 3D pseudo-GTs and stretched its training schedule two times.
Table~\ref{table:compare_estimation_network_3dpw} and ~\ref{table:compare_estimation_network_mupots} show that our 3DCrowdNet outperforms all existing methods on both 3DPW and MuPoTS.
In Table~\ref{table:compare_estimation_network_3dpw}, for the fair comparison with recent works~\cite{lin2021end,kocabas2021pare,lin2021mesh} that use 3DPW to train their networks, we additionally show our result when 3DCrowdNet is additionally trained on 3DPW.
Please note that 3DCrowdNet with its original 3D pseudo-GTs and stretched schedule produces a 50.1 3D error, much worse than our 46.1 3D error on 3DPW.
The tables show the power of our 3D pseudo-GTs, which elevate a state-of-the-art estimation network $g$ to a top-performing method.

\begin{table}[t]
\footnotesize
\centering
\setlength\tabcolsep{1.0pt}
\def\arraystretch{1.1}
\begin{tabular}{C{2.5cm}C{3.8cm}|C{1.8cm}}
\specialrule{.1em}{.05em}{.05em}
First stage & Second stage & 3D errors \\ \hline
\multirow{3}{*}{Annot. network $f$} & Annot. network $f$ & 48.21 / 75.55 \\
 & Fine-tuned annot. network $f$ &   47.93 / 75.13 \\
 & Est. network $g$ & \textbf{46.21 / 74.40} \\
\specialrule{.1em}{.05em}{.05em}
\end{tabular}
\vspace*{-2mm}
\caption{
Comparison of 3D errors (PA MPJPE) of various pipelines on 3DPW.
The left and right 3D errors of $g$ (PA MPJPE) are calculated on 3DPW and MuPoTS, respectively.
}
\label{table:justification_two_stage}
\end{table}

\begin{table*}[t]
\footnotesize
\centering
\setlength\tabcolsep{1.0pt}
\def\arraystretch{1.1}
\begin{tabular}{C{4.5cm}|C{4.2cm}|C{3.5cm}}
\specialrule{.1em}{.05em}{.05em}
Recipes &  Where to apply recipe & 3D errors of $g$ \\ \hline
\multirow{4}{*}{Train on 3DPW} & None & 50.82 / 75.63 \\
& Annotation network $f$ & 47.13 / 74.43 \\
& Estimation network $g$ &  48.33 / 74.84 \\ 
& Both $f$ and $g$ &  45.98 / 73.97 \\ \hline
\multirow{4}{*}{Initialize with 2D pose network} & None & 48.84 / 75.72 \\
& Annotation network $f$ & 46.99 / 74.58 \\
& Estimation network $g$ & 48.13 / 74.73 \\ 
& Both $f$ and $g$ & 45.98 / 73.97 \\ \hline
\multirow{4}{*}{Use VPoser and L2 reg. } & None & 48.31 / 75.70 \\
& Annotation network $f$ & 46.21 / 74.40 \\ 
& Estimation network $g$ & 48.13 / 75.72 \\ 
& Both $f$ and $g$ & 45.98 / 73.97 \\
 \specialrule{.1em}{.05em}{.05em}
\end{tabular}
\vspace*{-2mm}
\caption{
Comparison of 3D errors of estimation networks $g$, trained with different settings.
The left and right 3D errors of $g$ (PA MPJPE) are calculated on 3DPW and MuPoTS, respectively.
}
\label{table:where_to_apply_recipes}
\vspace{-.5em}
\end{table*}

\begin{table*}[t]
\footnotesize
\centering
\setlength\tabcolsep{1.0pt}
\def\arraystretch{1.1}
\vspace{-.5em}
\begin{tabular}{C{0.7cm}C{4.0cm}C{2.0cm}|C{3.5cm}C{2.2cm}}
\specialrule{.1em}{.05em}{.05em}
 \multicolumn{3}{c|}{Annotation network $f$} & \multicolumn{2}{c}{Estimation network $g$} \\
ID & Training sets & Unique images & Training sets & 3D errors \\ \hline
$f1$ & \multicolumn{1}{l}{H36M+MI+COCO} &  \multicolumn{1}{l|}{919K} & H36M+MI+[COCO]\textsubscript{$f1$} & 53.02 / 77.04 \\ 
$f2$ & \multicolumn{1}{l}{H36M+MI+COCO+MPII} & \multicolumn{1}{l|}{919K+29K} & H36M+MI+[COCO]\textsubscript{$f2$} & 53.23 / 77.05 \\
$f3$ & \multicolumn{1}{l}{H36M+MI+COCO+LSPET} & \multicolumn{1}{l|}{919K+9K} & H36M+MI+[COCO]\textsubscript{$f3$} & 54.14 / 77.53 \\ 
$f4$ & \multicolumn{1}{l}{H36M+MI+COCO+InstaVariety} & \multicolumn{1}{l|}{919K+2185K} & H36M+MI+[COCO]\textsubscript{$f4$} & 53.86 / 78.49 \\ 
$f5$ & \multicolumn{1}{l}{H36M+MI+COCO+3DPW} & \multicolumn{1}{l|}{919K+23K} & H36M+MI+[COCO]\textsubscript{$f5$} & \textbf{51.61} / \textbf{75.37} \\ 
\specialrule{.1em}{.05em}{.05em}
\end{tabular}
\vspace*{-2mm}
\caption{
Comparison of 3D errors of estimation networks $g$, trained with different 3D pseudo-GTs of COCO.
Notations are the same as Table~\ref{table:compare_annotation_pose2pose}.
}
\label{table:3dpw_itw_compare}
\end{table*}

\begin{table*}[t]
\footnotesize
\centering
\setlength\tabcolsep{1.0pt}
\def\arraystretch{1.1}
\begin{tabular}{C{0.7cm}C{6.2cm}|C{3.5cm}C{3.1cm}}
\specialrule{.1em}{.05em}{.05em}
\multicolumn{2}{c|}{Annotation network $f$} & \multicolumn{2}{c}{Estimation network $g$} \\
ID & Training sets & Training sets & 3D errors \\ \hline
$f1$ & \multicolumn{1}{l|}{H36M+MI+COCO} &  H36M+MI+[COCO]\textsubscript{$f1$} & 53.02  (18.7/19.4/38.1) \\ 
$f2$ & \multicolumn{1}{l|}{H36M+MI+COCO+3DPW without 3D GTs} & H36M+MI+[COCO]\textsubscript{$f2$}  &  53.66 (18.8/19.6/38.6) \\
$f3$ & \multicolumn{1}{l|}{H36M+MI+COCO+3DPW} & H36M+MI+[COCO]\textsubscript{$f3$} & \textbf{51.61 (18.4/19.2/36.9)} \\ \specialrule{.1em}{.05em}{.05em}
\end{tabular}
\vspace*{-2mm}
\caption{
Comparison of 3D errors of estimation networks $g$, trained with different 3D pseudo-GTs of COCO.
Notations are the same as Table~\ref{table:compare_annotation_various}.
}
\label{table:3dpw_wo_3d_gts}
\vspace{-1em}
\end{table*}

\begin{table*}[!htbp]
\footnotesize
\centering
\setlength\tabcolsep{1.0pt}
\def\arraystretch{1.1}
\begin{tabular}{C{0.5cm}C{2.5cm}C{4.0cm}|C{3.5cm}C{3.0cm}}
\specialrule{.1em}{.05em}{.05em}
\multicolumn{3}{c|}{Annotation network $f$} & \multicolumn{2}{c}{Estimation network $g$} \\
ID & Initialization &  Training sets &  Training sets & 3D errors \\ \hline
$f1$ & ImageNet cls.~\cite{he2016deep} &  \multirow{3}{*}{H36M+MI+COCO+3DPW} &   H36M+MI+[COCO]\textsubscript{$f1$} & 51.61 (18.4/19.2/36.9) \\
$f2$ & 3D pose~\cite{kolotouros2019learning} & &  H36M+MI+[COCO]\textsubscript{$f2$} & 51.62 (18.4/19.0/37.0)  \\
$f3$ & 2D pose~\cite{xiao2018simple} &   & H36M+MI+[COCO]\textsubscript{$f3$} & \textbf{47.52 (17.4/18.4/33.0)}  \\ \specialrule{.1em}{.05em}{.05em}
\end{tabular}
\vspace*{-2mm}
\caption{
Comparison of 3D errors of estimation networks $g$, trained with different 3D pseudo-GTs of COCO.
Notations are the same as Table~\ref{table:compare_annotation_various}.
}
\label{table:2d_pose_init_annotation_network}
\vspace{-1.5em}
\end{table*}

\subsection{Ablation study}
For all the ablation studies, our annotation network $f$ produces 3D pseudo-GTs of COCO.
Then, the 3D pseudo-GTs of COCO in addition to H36M, MI, and optionally 3DPW are used to train the estimation network $g$.
We use Pose2Pose~\cite{moon2022hand4whole} for our estimation network $g$.

\noindent\textbf{Justification of using separated networks in our two-stage framework.}
As shown in Fig.~\ref{fig:overall_pipeline}, our framework for the 3D human mesh estimation in the wild consists of two stages, of which the first stage is obtaining 3D pseudo-GTs with annotation network $f$, and the second stage is training estimation network $g$ with the 3D pseudo-GTs of the first stage.
The annotation network $f$ and estimation network $g$ are separated.
To justify using separated networks in our two-stage framework, we compare two variants with our setting in Table~\ref{table:justification_two_stage}.
The first variant is testing the annotation network $f$ on 3DPW.
Although the purpose of the annotation network is to obtain 3D pseudo-GTs of ITW datasets, we can use it as an estimation network and test it on 3DPW.
The second variant measures the 3D error using the annotation network $f$ after fine-tuning it with the 3D pseudo-GTs.
This setting uses 3D pseudo-GTs from the annotation network $f$ (the first variant) for the fine-tuning; however, it uses one network instead of the separated two networks like ours.
Finally, our estimation network $g$ is trained with the 3D pseudo-GTs, where the 3D pseudo-GTs are from the annotation network $f$ (the first variant).
For a fair comparison, all networks in the table have the same network architecture of Pose2Pose~\cite{moon2022hand4whole}.
The table shows that our estimation network $g$ achieves the lowest error, which justifies using separated networks in our two-stage framework.
The reason for the better performance of our setting is that it is \emph{fully supervised} with 3D targets (\textit{i.e.}, 3D pseudo-GTs) from the start of the training.
On the other hand, the two variants are \emph{weakly supervised} with 2D targets (\textit{i.e.}, 2D GT poses) from the start of the training, although the second variant is fine-tuned with 3D targets later.
The weak supervision at the start of the training makes the two variants converge to sub-optimal points compared to the estimation network $g$.
All networks in the table are trained on H36M+MI+MSCOCO+3DPW.
In addition, ResNet~\cite{he2016deep} of them are initialized with pre-trained 2D pose estimation network~\cite{xiao2018simple}.

\noindent\textbf{Applying the recipes to estimation networks \boldmath{$g$}.}
Table~\ref{table:where_to_apply_recipes} shows that applying our recipes to annotation network $f$ improves the 3D errors on both benchmarks much more than applying them to estimation networks $g$.
Applying the recipes to both annotation network $f$ and estimation network $g$ performs the best; however, the performance improvement is limited compared to the improvement brought by applying the recipes only to annotation network $f$.
For example, applying each recipe to annotation network $f$ brings 3.69, 1.85, and 2.1 3D error improvement, respectively, while applying additionally to both $f$ and $g$ brings 1.15, 1.01, and 0.23 3D error improvement.
This shows that our recipes are specially designed for the annotation networks $f$ to obtain beneficial 3D pseudo-GTs.
The reason for the small effect when the recipes are applied to the estimation networks $g$ is that the estimation networks $g$ are trained with 3D pseudo-GTs, while annotation networks $f$ are trained with 2D GTs of ITW datasets without 3D evidence.
The absence of 3D evidence when training annotation networks $f$ results in severe ambiguities, which can be cured by our recipes.
On the other hand, as estimation networks $g$ are fully supervised with 3D pseudo-GTs, they suffer less from ambiguities.
For each recipe, \emph{None} represents both annotation and estimation networks are trained with the remaining other two recipes.

\noindent\textbf{Effect of training annotation network \boldmath{$f$} on 3DPW.}
Table~\ref{table:3dpw_itw_compare} shows how 3DPW changes the 3D pseudo-GTs compared to other ITW datasets, such as MPII, LSPET, and InstaVariety~\cite{kanazawa2019learning}.
As the table shows, adding other ITW datasets does not obtain the performance gain of $g$ compared to H36M+MI+COCO.
This is because adding ITW datasets does not contribute to relieving the depth ambiguity as they provide only 2D GTs.
On the other hand, 3DPW provides 3D GTs, largely helpful to alleviate the depth ambiguity.
Importantly, the 3D errors of $g$ on MuPoTS decrease as well, which implies that using 3DPW as an additional training set is beneficial for multiple 3D ITW benchmarks.

It is noticeable that InstaVariety has 95 times more images than 3DPW, while much less helpful for the beneficial 3D pseudo-GTs.
This tells us that for the 3D pseudo-GTs of ITW datasets, the existence of 3D GTs is much more important than a large number of 2D GTs and rich appearance distribution from ITW datasets.
It suggests a different research direction compared to recent representation learning methods~\cite{he2020momentum,grill2020bootstrap,henaff2021efficient} as they suggest that collecting large-scale unlabeled images can boost the image classification performance a lot.
Our analysis is consistent with Table~\ref{table:3dpw_wo_3d_gts}.
The table shows that when we only use 2D GTs of 3DPW without 3D GTs, the quality of 3D pseudo-GTs does not change much, which leads to similar 3D errors of $g$ compared to H36M+MI+COCO.
The result shows that the performance gain from using 3DPW is not from images of 3DPW, but from 3D GTs of 3DPW.
In particular, most of the performance gain is from the $z$-axis, which shows the effectiveness of using 3DPW to resolve the depth ambiguity.

\noindent\textbf{Effect of initializing annotation network \boldmath{$f$} with a pre-trained 2D pose network.}
Table~\ref{table:2d_pose_init_annotation_network} shows that initializing annotation network $f$ with a pre-trained 2D pose network~\cite{xiao2018simple} produces more beneficial 3D pseudo-GTs, which result in lower 3D errors of $g$ compared to the conventional ImageNet classification pre-training~\cite{he2016deep}.
This is because initializing with the pre-trained 2D pose network makes the annotation network $f$ extract useful human articulation features from images at the early stage of the training.
Therefore, better initialization results in a better convergence point, which alleviates the sub-optimality of weak supervision.
Interestingly, the $z$-axis error decreases much, while the errors of $x$- and $y$-axis remain similar.
This indicates that the proposed initialization does not simply result in better 2D pose estimation ability, but helps our annotation network $f$ to produce more beneficial 3D pseudo-GTs.
We further compare our initialization with the initialization of EFT~\cite{joo2021eft}, which initializes their network with pre-trained 3D pose estimation network~\cite{kolotouros2019learning}.
We observed that initializing the network with pre-trained 3D pose estimation network~\cite{kolotouros2019learning} produces almost the same results as the ImageNet counterpart and is largely beaten by our 2D-based initialization.
We think this is because the pre-trained 3D pose network~\cite{kolotouros2019learning} is already converged to produce lower quality 3D pseudo-GTs than ours.

\section{Related works}

SMPLify~\cite{bogo2016keep} and SMPLify-X~\cite{pavlakos2019expressive} are iterative fitting frameworks, which iteratively fit SMPL parameters to target 2D pose by minimizing energy functions.
Using them to 2D GT pose of ITW datasets, researchers~\cite{moon2020i2l,choi2020p2m,lin2021end,lin2021mesh} obtained 3D pseudo-GTs.
Recently, several annotation networks are introduced.
SPIN~\cite{kolotouros2019learning} predicts SMPL parameters using a network and iteratively fits~\cite{bogo2016keep} the predicted parameters to 2D GT pose.
Their final 3D pseudo-GT of each sample is obtained by selecting one with smaller SMPLify loss~\cite{bogo2016keep} between their fit and prepared initial 3D pseudo-GT.
The initial 3D pseudo-GTs are prepared before training their network by running SMPLify~\cite{bogo2016keep} to 2D GT pose.
The final 3D pseudo-GTs are used to train an HMR~\cite{kanazawa2018end} regressor. 
EFT~\cite{joo2021eft} fine-tunes the pre-trained SPIN to the 2D GT pose of each sample, and the outputs of the last fine-tuning iteration become the 3D pseudo-GT of the sample.
Both SPIN and EFT require initial 3D pseudo-GTs from SMPLify~\cite{bogo2016keep} to train their networks.
On the other hand, NeuralAnnot~\cite{moon2022neuralannot} is weakly supervised with 2D GT pose without requiring initial 3D pseudo-GTs.
Compare to them, our annotation network produces more beneficial 3D pseudo-GTs, which results in much lower 3D errors of the estimation networks (Table~\ref{table:compare_annotation_pose2pose} and ~\ref{table:compare_annotation_various}).
Table~\ref{table:compare_annotation_network_novelty} shows differences between our annotation networks and the above ones.

\section{Conclusion}
We introduce three recipes to obtain highly beneficial 3D pseudo-GTs of ITW datasets for the 3D human mesh estimation in the wild.
Experimental results show that simply re-training state-of-the-art networks with our 3D pseudo-GTs elevates their performance to the next level.
In addition, we show our 3D pseudo-GTs are much more beneficial than previous ones.
In closing, we hope the community to have more remarks on the importance of 3D pseudo-GTs.

\clearpage

\begin{center}
\textbf{\large Supplementary Material for \\ ``Three Recipes for Better 3D Pseudo-GTs of \\3D Human Mesh Estimation in the Wild"}
\end{center}

\setcounter{section}{0}
\setcounter{table}{0}
\setcounter{figure}{0}

\renewcommand{\thesection}{\Alph{section}}   
\renewcommand{\thetable}{\Alph{table}}   
\renewcommand{\thefigure}{\Alph{figure}}

In this supplementary material, we provide more experiments, discussions, and other details that could not be included in the main text due to the lack of pages.
The contents are summarized below:
\begin{enumerate}[nosep, label=\Alph*.] 
    \item Effectiveness of a combination of VPoser~\cite{pavlakos2019expressive} and L2 regularizer
    \item Qualitative comparisons
    \item Implementation details
    \item Limitations
\end{enumerate}

\section{Effectiveness of a combination of VPoser and L2 regularizer}

Table~\ref{table:vposer_l2_reg} shows the effectiveness of 1) usage of VPoser~\cite{pavlakos2019expressive} and 2) weight of L2 regularizer during the training of the annotation network $f$.
The combination of VPoser and L2 regularizer is the third recipe, introduced in Section 2.2 of the main manuscript.
Regardless of the usage of VPoser, setting the non-zero weight of the L2 regularizer produces lower 3D errors of $g$.
This indicates that despite its simplicity, the L2 regularizer helps to prevent anatomically implausible 3D meshes and produce beneficial 3D pseudo-GTs.
In addition, using VPoser achieves lower 3D errors of $g$ compared to not using it.
This is also because VPoser can effectively limit the 3D mesh to anatomically plausible space.
Training sets of all annotation networks $f$ in the table are H36M+MI+MSCOCO+3DPW.
The ResNet backbone~\cite{he2016deep} of all annotation networks in the table are initialized with ResNet, pre-trianed on ImageNet~\cite{russakovsky2015imagenet} classification dataset.

\begin{figure*}[t]
\begin{center}
\includegraphics[width=\linewidth]{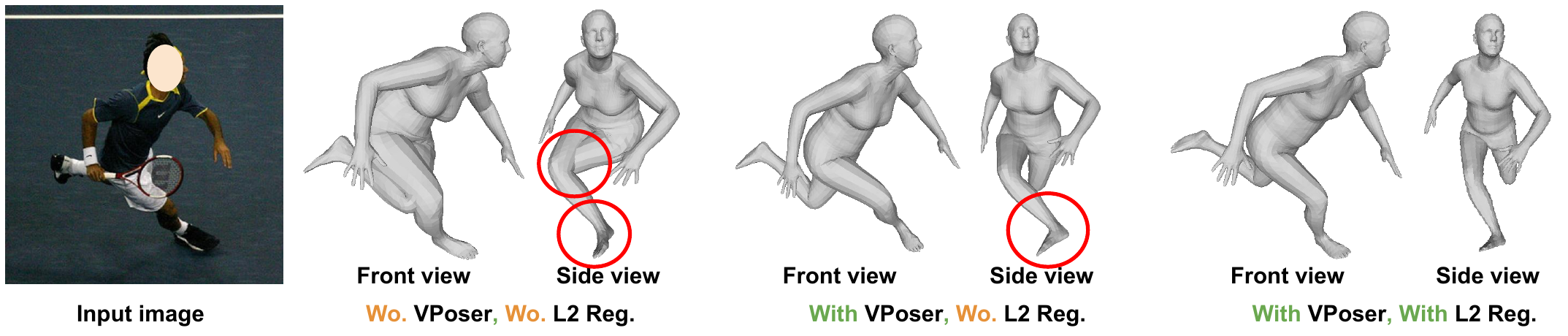}
\end{center}
   \caption{
   Visual comparison between 3D pseudo-GTs, obtained from annotation networks whose IDs are $f1\text{-}1$, $f2\text{-}1$, and $f2\text{-}3$ of Table~\ref{table:vposer_l2_reg}.
   Wrong parts are highlighted.
   }
\label{fig:vposer_l2_reg}
\end{figure*}

Fig.~\ref{fig:vposer_l2_reg} shows the effectiveness of 1) using VPoser and 2) applying L2 regularizer.
Without VPoser and L2 regularizer, the 3D pseudo-GT has an anatomically implausible 3D mesh although its 2D pose is fit to the image.
Using VPoser makes the 3D pseudo-GT anatomically plausible; however, it still produces the wrong 3D mesh.
The right leg is too much bent to the left side.
Finally, using both VPoser and L2 regularizer makes the 3D pseudo-GT anatomically plausible and correct.
In particular, additionally using the L2 regularizer enforces the 3D mesh in the latent space of VPoser.

\begin{table*}[t]
\small
\centering
\setlength\tabcolsep{1.0pt}
\def\arraystretch{1.1}
\begin{tabular}{C{1.0cm}C{2.0cm}C{3.0cm}|C{4.5cm}C{3.0cm}}
\specialrule{.1em}{.05em}{.05em}
\multicolumn{3}{c|}{Annotation network $f$} & \multicolumn{2}{c}{Estimation network $g$} \\
ID & Use VPoser & L2 reg. weight & Training sets & 3D errors \\ \hline
$f1\text{-}1$ & \xmark &  0.0  &   H36M+MI+[MSCOCO]\textsubscript{$f1\text{-}1$} & 65.98 \\
$f1\text{-}2$ & \xmark &  $10^{-1}$  &   H36M+MI+[MSCOCO]\textsubscript{$f1\text{-}2$} & 79.07 \\
$f1\text{-}3$ & \xmark &  $10^{-2}$  &   H36M+MI+[MSCOCO]\textsubscript{$f1\text{-}3$} & 64.25 \\
$f1\text{-}4$ & \xmark &  $10^{-3}$  &   H36M+MI+[MSCOCO]\textsubscript{$f1\text{-}4$} & 64.04 \\
$f1\text{-}5$ & \xmark & $10^{-4}$ &
H36M+MI+[MSCOCO]\textsubscript{$f1\text{-}5$} & 64.67 \\
$f1\text{-}6$ & \xmark & $10^{-5}$ &H36M+MI+[MSCOCO]\textsubscript{$f1\text{-}6$} &  64.39 \\ \hline
$f2\text{-}1$ & \cmark &  0.0  &   H36M+MI+[MSCOCO]\textsubscript{$f2\text{-}1$} & 56.57 \\
$f2\text{-}2$ & \cmark &  $10^{-1}$  &   H36M+MI+[MSCOCO]\textsubscript{$f2\text{-}2$} & 59.22 \\
$f2\text{-}3$ & \cmark &  $10^{-2}$  &   H36M+MI+[MSCOCO]\textsubscript{$f2\text{-}3$} & \textbf{51.61} \\
$f2\text{-}4$ & \cmark &  $10^{-3}$  &   H36M+MI+[MSCOCO]\textsubscript{$f2\text{-}4$} &  53.27\\
$f2\text{-}5$ & \cmark & $10^{-4}$ &
H36M+MI+[MSCOCO]\textsubscript{$f2\text{-}5$} & 55.06 \\
$f2\text{-}6$ & \cmark & $10^{-5}$ &H36M+MI+[MSCOCO]\textsubscript{$f2\text{-}6$} &  56.03 \\
\specialrule{.1em}{.05em}{.05em}
\end{tabular}
\vspace{.5em}
\caption{
Comparison of 3D errors of estimation networks $g$, trained with different 3D pseudo-GTs of MSCOCO.
The subscript at the square brackets denotes a method to obtain the 3D pseudo-GTs.
The 3D errors of $g$ (PA MPJPE) are calculated on 3DPW.
}
\label{table:vposer_l2_reg}
\end{table*}

\begin{figure*}[t]
\begin{center}
\includegraphics[width=\linewidth]{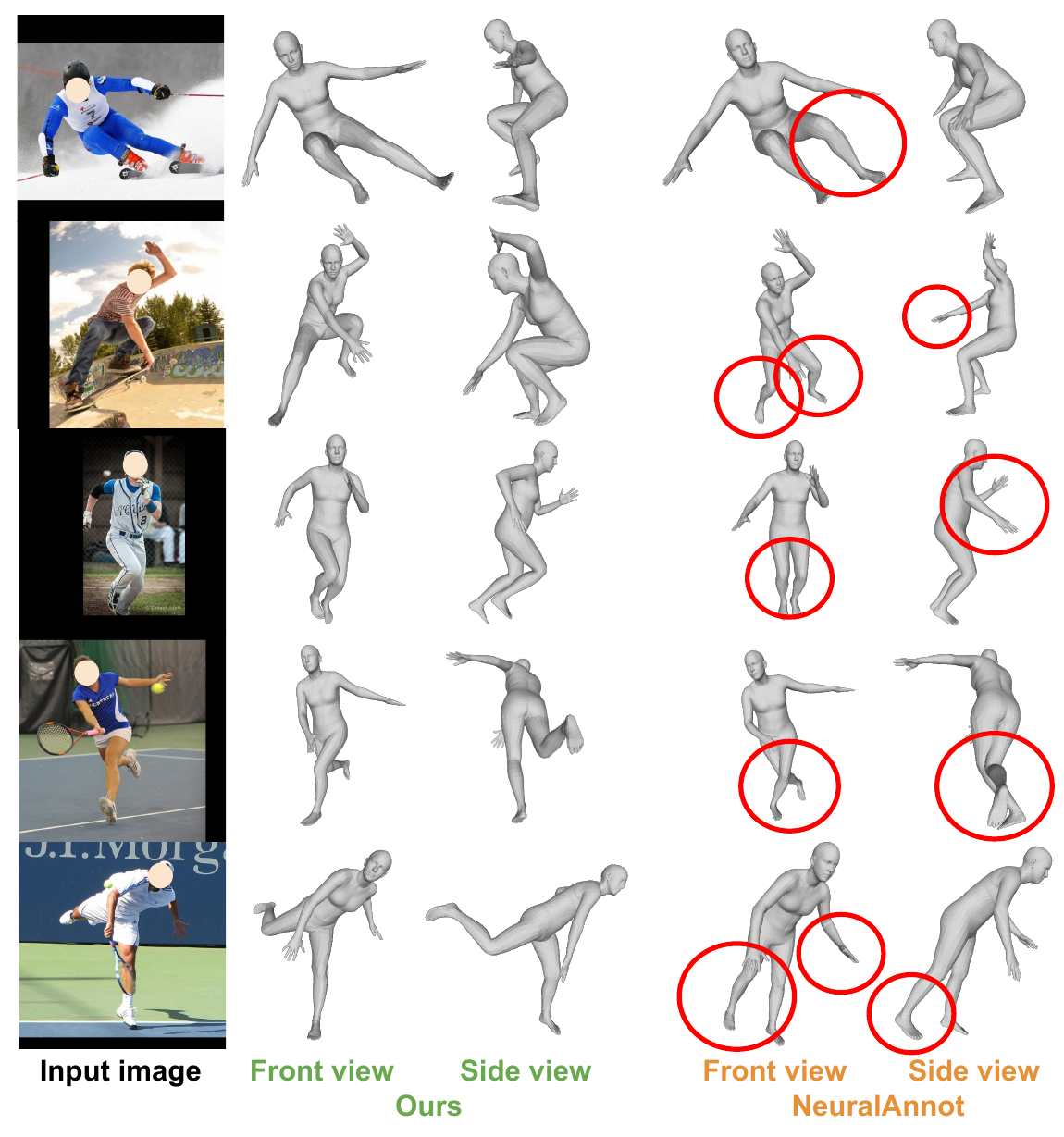}
\end{center}
   \caption{
   Visual comparison between 3D pseudo-GTs of MSCOCO from ours and NeuralAnnot~\cite{moon2022neuralannot}.
   Wrong parts are highlighted.
   }
\label{fig:qualitative_comparison_suppl}
\end{figure*}

\begin{figure*}[t]
\begin{center}
\includegraphics[width=\linewidth]{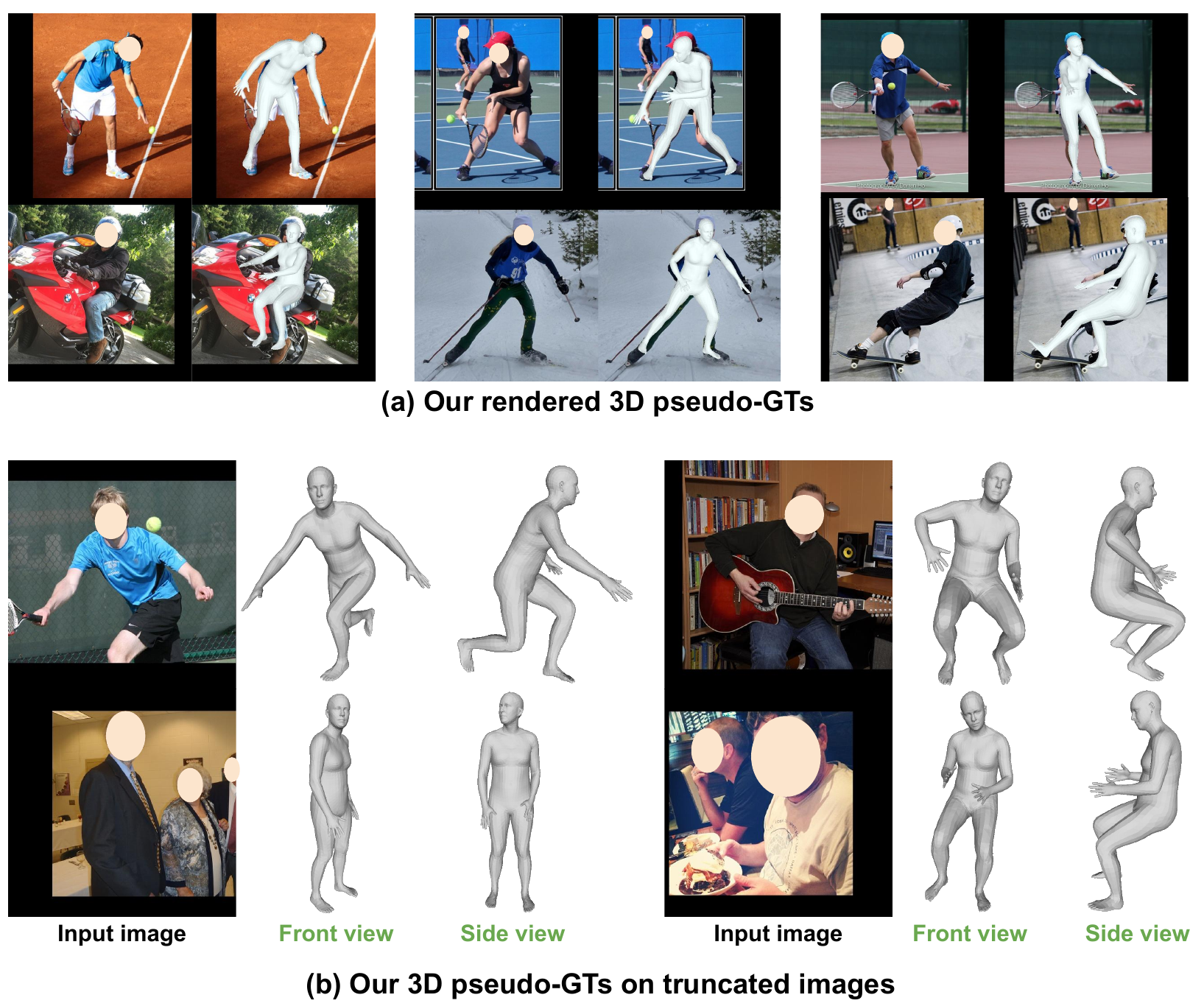}
\end{center}
   \caption{
   (a) Our rendered 3D pseudo-GTs on images of MSCOCO.
   (b) Our 3D pseudo-GTs on truncated images of MSCOCO.
   }
\label{fig:qualitative_results}
\end{figure*}

\section{Qualitative comparisons}

Fig.~\ref{fig:qualitative_comparison_suppl} shows qualitative comparisons between 3D pseudo-GTs from our annotation network $f$ and NeuralAnnot~\cite{moon2022neuralannot}.
The comparisons show that our 3D pseudo-GTs are more accurate than those of NeuralAnnot.
The results from the first row to the fourth row show that ours are more robust to the depth ambiguity.
For example, in the fourth row example, both have almost the same 2D position of the right knee.
However, our 3D position of the right knee is more accurate as it does not penetrate inside of the left leg.

Fig.~\ref{fig:qualitative_results} shows (a) rendered 3D pseudo-GTs on various images of MSCOCO and (b) 3D pseudo-GTs on truncated images of MSCOCO.
The rendered results show that our 3D pseudo-GTs are well-aligned with the image.
In addition, ours produce robust 3D pseudo-GTs on severely truncated images by utilizing strong contextual information of image features.

\section{Implementation details of annotation network $f$}

As described in Section 2.1 of the main manuscript, our annotation network $f$ is based on Pose2Pose network~\cite{moon2022hand4whole}.
Hence, most of the details follow theirs.
PyTorch~\cite{paszke2017automatic} is used for implementation. 
For the training, we use Adam optimizer~\cite{kingma2014adam} with a mini-batch size of 192.
Data augmentations, including scaling, rotation, random horizontal flip, and color jittering, are performed during the training.
The initial learning rate is set to $10^{-4}$ and reduced by a factor of 10 at the \nth{11} and \nth{13} epoch.
We train our annotation network $f$ for 15 epochs.
A single NVIDIA A100 GPU is used for the experiments, where it takes 6 hours to train our annotation network $f$.
We modified the Pose2Pose network to predict the latent code of VPoser instead of SMPL pose parameters.
The predicted VPoser latent code is passed to the decoder of VPoser, which outputs the SMPL pose parameter.
The L2 regularizer is applied to the predicted latent code of VPoser and SMPL shape parameter, where its weight is determined to $10^{-2}$ following Table~\ref{table:vposer_l2_reg}.
A neutral gender SMPL model is used for the experiments.
All other details are available in the codes of Pose2Pose~\cite{moon2022hand4whole}\footnote{\url{https://github.com/mks0601/Hand4Whole_RELEASE/tree/Pose2Pose}}.

\section{Limitations}

Although our annotation network $f$ produces much more beneficial 3D pseudo-GTs than previous attempts, our 3D pseudo-GTs still contain some errors in nature.
This could be addressed by collecting more ITW 3D datasets, such as 3DPW, as we made our 3D pseudo-GTs more beneficial by utilizing 3DPW to train annotation network $f$.
Collecting ITW 3D datasets is challenging; however, we believe it is worthwhile considering its usefulness.
In particular, Table 8 of the main manuscript shows that 3DPW is much more helpful than existing large-scale ITW 2D datasets, such as InstaVariety, despite the small scale of 3DPW.
We observed from an additional study that using 50\% and 10\% of 3DPW when training the annotation network $f$ decreases 3D error of the estimation network $g$ only 4\% and 7\%, respectively.
As such analysis shows that even a small amount of ITW 3D datasets are helpful, it relieves a concern on collection costs of ITW 3D datasets.

\clearpage

{\small
\bibliographystyle{ieee_fullname}
\bibliography{bib}

\begin{thebibliography}{10}\itemsep=-1pt

\bibitem{andriluka20142d}
Mykhaylo Andriluka, Leonid Pishchulin, Peter Gehler, and Bernt Schiele.
\newblock {2D} human pose estimation: New benchmark and state of the art
  analysis.
\newblock In {\em CVPR}, 2014.

\bibitem{bilen2016weakly}
Hakan Bilen and Andrea Vedaldi.
\newblock Weakly supervised deep detection networks.
\newblock In {\em CVPR}, 2016.

\bibitem{bogo2016keep}
Federica Bogo, Angjoo Kanazawa, Christoph Lassner, Peter Gehler, Javier Romero,
  and Michael~J Black.
\newblock {Keep it SMPL}: Automatic estimation of {3D} human pose and shape
  from a single image.
\newblock In {\em ECCV}, 2016.

\bibitem{choe2022evaluation}
Junsuk Choe, Seong~Joon Oh, Sanghyuk Chun, Seungho Lee, Zeynep Akata, and
  Hyunjung Shim.
\newblock Evaluation for weakly supervised object localization: Protocol,
  metrics, and datasets.
\newblock {\em TPAMI}, 2022.

\bibitem{choe2020wsoleval}
Junsuk Choe, Seong~Joon Oh, Seungho Lee, Sanghyuk Chun, Zeynep Akata, and
  Hyunjung Shim.
\newblock Evaluating weakly supervised object localization methods right.
\newblock In {\em CVPR}, 2020.

\bibitem{choi2021beyond}
Hongsuk Choi, Gyeongsik Moon, Ju~Yong Chang, and Kyoung~Mu Lee.
\newblock Beyond static features for temporally consistent {3D} human pose and
  shape from a video.
\newblock In {\em CVPR}, 2021.

\bibitem{choi2020p2m}
Hongsuk Choi, Gyeongsik Moon, and Kyoung~Mu Lee.
\newblock {Pose2Mesh}: Graph convolutional network for {3D} human pose and mesh
  recovery from a {2D} human pose.
\newblock In {\em ECCV}, 2020.

\bibitem{choi2022learning}
Hongsuk Choi, Gyeongsik Moon, JoonKyu Park, and Kyoung~Mu Lee.
\newblock Learning to estimate robust {3D} human mesh from in-the-wild crowded
  scenes.
\newblock In {\em CVPR}, 2022.

\bibitem{durand2017wildcat}
Thibaut Durand, Taylor Mordan, Nicolas Thome, and Matthieu Cord.
\newblock {WILDCAT}: Weakly supervised learning of deep convnets for image
  classification, pointwise localization and segmentation.
\newblock In {\em CVPR}, 2017.

\bibitem{grill2020bootstrap}
Jean-Bastien Grill, Florian Strub, Florent Altch{\'e}, Corentin Tallec, Pierre
  Richemond, Elena Buchatskaya, Carl Doersch, Bernardo Avila~Pires, Zhaohan
  Guo, Mohammad Gheshlaghi~Azar, et~al.
\newblock Bootstrap your own latent-a new approach to self-supervised learning.
\newblock In {\em NeurIPS}, 2020.

\bibitem{he2020momentum}
Kaiming He, Haoqi Fan, Yuxin Wu, Saining Xie, and Ross Girshick.
\newblock Momentum contrast for unsupervised visual representation learning.
\newblock In {\em CVPR}, 2020.

\bibitem{he2016deep}
Kaiming He, Xiangyu Zhang, Shaoqing Ren, and Jian Sun.
\newblock Deep residual learning for image recognition.
\newblock In {\em CVPR}, 2016.

\bibitem{henaff2021efficient}
Olivier~J H{\'e}naff, Skanda Koppula, Jean-Baptiste Alayrac, Aaron van~den
  Oord, Oriol Vinyals, and Jo{\~a}o Carreira.
\newblock Efficient visual pretraining with contrastive detection.
\newblock In {\em ICCV}, 2021.

\bibitem{ionescu2014human3}
Catalin Ionescu, Dragos Papava, Vlad Olaru, and Cristian Sminchisescu.
\newblock {Human3.6M}: Large scale datasets and predictive methods for {3D}
  human sensing in natural environments.
\newblock {\em TPAMI}, 2014.

\bibitem{jiang2020coherent}
Wen Jiang, Nikos Kolotouros, Georgios Pavlakos, Xiaowei Zhou, and Kostas
  Daniilidis.
\newblock Coherent reconstruction of multiple humans from a single image.
\newblock In {\em CVPR}, 2020.

\bibitem{johnson2011learning}
Sam Johnson and Mark Everingham.
\newblock Learning effective human pose estimation from inaccurate annotation.
\newblock In {\em CVPR}, 2011.

\bibitem{joo2015panoptic}
Hanbyul Joo, Hao Liu, Lei Tan, Lin Gui, Bart Nabbe, Iain Matthews, Takeo
  Kanade, Shohei Nobuhara, and Yaser Sheikh.
\newblock {Panoptic Studio}: A massively multiview system for social motion
  capture.
\newblock In {\em ICCV}, 2015.

\bibitem{joo2021eft}
Hanbyul Joo, Natalia Neverova, and Andrea Vedaldi.
\newblock Exemplar fine-tuning for {3D} human pose fitting towards in-the-wild
  {3D} human pose estimation.
\newblock In {\em 3DV}, 2021.

\bibitem{kanazawa2018end}
Angjoo Kanazawa, Michael~J Black, David~W Jacobs, and Jitendra Malik.
\newblock End-to-end recovery of human shape and pose.
\newblock In {\em CVPR}, 2018.

\bibitem{kanazawa2019learning}
Angjoo Kanazawa, Jason~Y Zhang, Panna Felsen, and Jitendra Malik.
\newblock Learning {3D} human dynamics from video.
\newblock In {\em CVPR}, 2019.

\bibitem{kingma2014adam}
Diederik~P Kingma and Jimmy Ba.
\newblock Adam: A method for stochastic optimization.
\newblock In {\em ICLR}, 2014.

\bibitem{kocabas2020vibe}
Muhammed Kocabas, Nikos Athanasiou, and Michael~J Black.
\newblock {VIBE}: Video inference for human body pose and shape estimation.
\newblock In {\em CVPR}, 2020.

\bibitem{kocabas2021pare}
Muhammed Kocabas, Chun-Hao~P Huang, Otmar Hilliges, and Michael~J Black.
\newblock {PARE}: Part attention regressor for {3D} human body estimation.
\newblock In {\em ICCV}, 2021.

\bibitem{kolotouros2019learning}
Nikos Kolotouros, Georgios Pavlakos, Michael~J Black, and Kostas Daniilidis.
\newblock Learning to reconstruct {3D} human pose and shape via model-fitting
  in the loop.
\newblock In {\em ICCV}, 2019.

\bibitem{li2021hybrik}
Jiefeng Li, Chao Xu, Zhicun Chen, Siyuan Bian, Lixin Yang, and Cewu Lu.
\newblock {HybrIK}: A hybrid analytical-neural inverse kinematics solution for
  {3D} human pose and shape estimation.
\newblock In {\em CVPR}, 2021.

\bibitem{li2022cliff}
Zhihao Li, Jianzhuang Liu, Zhensong Zhang, Songcen Xu, and Youliang Yan.
\newblock {CLIFF}: Carrying location information in full frames into human pose
  and shape estimation.
\newblock In {\em ECCV}, 2022.

\bibitem{lin2021end}
Kevin Lin, Lijuan Wang, and Zicheng Liu.
\newblock End-to-end human pose and mesh reconstruction with transformers.
\newblock In {\em CVPR}, 2021.

\bibitem{lin2021mesh}
Kevin Lin, Lijuan Wang, and Zicheng Liu.
\newblock Mesh graphormer.
\newblock In {\em ICCV}, 2021.

\bibitem{lin2014microsoft}
Tsung-Yi Lin, Michael Maire, Serge Belongie, James Hays, Pietro Perona, Deva
  Ramanan, Piotr Doll{\'a}r, and C~Lawrence Zitnick.
\newblock {Microsoft COCO}: Common objects in context.
\newblock In {\em ECCV}, 2014.

\bibitem{loper2015smpl}
Matthew Loper, Naureen Mahmood, Javier Romero, Gerard Pons-Moll, and Michael~J
  Black.
\newblock {SMPL}: A skinned multi-person linear model.
\newblock {\em ACM TOG}, 2015.

\bibitem{mahmood2019amass}
Naureen Mahmood, Nima Ghorbani, Nikolaus~F Troje, Gerard Pons-Moll, and
  Michael~J Black.
\newblock {AMASS}: Archive of motion capture as surface shapes.
\newblock In {\em ICCV}, 2019.

\bibitem{mehta2017monocular}
Dushyant Mehta, Helge Rhodin, Dan Casas, Pascal Fua, Oleksandr Sotnychenko,
  Weipeng Xu, and Christian Theobalt.
\newblock Monocular {3D} human pose estimation in the wild using improved {CNN}
  supervision.
\newblock In {\em 3DV}, 2017.

\bibitem{mehta2018single}
Dushyant Mehta, Oleksandr Sotnychenko, Franziska Mueller, Weipeng Xu, Srinath
  Sridhar, Gerard Pons-Moll, and Christian Theobalt.
\newblock Single-shot multi-person {3D} pose estimation from monocular {RGB}.
\newblock In {\em 3DV}, 2018.

\bibitem{moon2022hand4whole}
Gyeongsik Moon, Hongsuk Choi, and Kyoung~Mu Lee.
\newblock Accurate {3D} hand pose estimation for whole-body {3D} human mesh
  estimation.
\newblock In {\em CVPRW}, 2022.

\bibitem{moon2022neuralannot}
Gyeongsik Moon, Hongsuk Choi, and Kyoung~Mu Lee.
\newblock {NeuralAnnot}: Neural annotator for {3D} human mesh training sets.
\newblock In {\em CVPRW}, 2022.

\bibitem{moon2020i2l}
Gyeongsik Moon and Kyoung~Mu Lee.
\newblock {I2L-MeshNet}: {Image-to-Lixel} prediction network for accurate {3D}
  human pose and mesh estimation from a single {RGB} image.
\newblock In {\em ECCV}, 2020.

\bibitem{moon20223d}
Gyeongsik Moon, Hyeongjin Nam, Takaaki Shiratori, and Kyoung~Mu Lee.
\newblock {3D} clothed human reconstruction in the wild.
\newblock In {\em ECCV}, 2022.

\bibitem{moon2020interhand2}
Gyeongsik Moon, Shoou-I Yu, He Wen, Takaaki Shiratori, and Kyoung~Mu Lee.
\newblock {InterHand2.6M}: A dataset and baseline for {3D} interacting hand
  pose estimation from a single {RGB} image.
\newblock In {\em ECCV}, 2020.

\bibitem{paszke2017automatic}
Adam Paszke, Sam Gross, Soumith Chintala, Gregory Chanan, Edward Yang, Zachary
  DeVito, Zeming Lin, Alban Desmaison, Luca Antiga, and Adam Lerer.
\newblock Automatic differentiation in pytorch.
\newblock 2017.

\bibitem{pavlakos2019expressive}
Georgios Pavlakos, Vasileios Choutas, Nima Ghorbani, Timo Bolkart, Ahmed~AA
  Osman, Dimitrios Tzionas, and Michael~J Black.
\newblock Expressive body capture: {3D} hands, face, and body from a single
  image.
\newblock In {\em CVPR}, 2019.

\bibitem{russakovsky2015imagenet}
Olga Russakovsky, Jia Deng, Hao Su, Jonathan Krause, Sanjeev Satheesh, Sean Ma,
  Zhiheng Huang, Andrej Karpathy, Aditya Khosla, Michael Bernstein, et~al.
\newblock {ImageNet} large scale visual recognition challenge.
\newblock {\em IJCV}, 2015.

\bibitem{ROMP}
Yu Sun, Qian Bao, Wu Liu, Yili Fu, Black Michael~J., and Tao Mei.
\newblock Monocular, one-stage, regression of multiple {3D} people.
\newblock In {\em ICCV}, 2021.

\bibitem{tang2018pcl}
Peng Tang, Xinggang Wang, Song Bai, Wei Shen, Xiang Bai, Wenyu Liu, and Alan
  Yuille.
\newblock {PCL}: Proposal cluster learning for weakly supervised object
  detection.
\newblock {\em TPAMI}, 2018.

\bibitem{von2018recovering}
Timo von Marcard, Roberto Henschel, Michael~J Black, Bodo Rosenhahn, and Gerard
  Pons-Moll.
\newblock Recovering accurate {3D} human pose in the wild using {IMUs} and a
  moving camera.
\newblock In {\em ECCV}, 2018.

\bibitem{wan2019c}
Fang Wan, Chang Liu, Wei Ke, Xiangyang Ji, Jianbin Jiao, and Qixiang Ye.
\newblock {C-MIL}: Continuation multiple instance learning for weakly
  supervised object detection.
\newblock In {\em CVPR}, 2019.

\bibitem{xiao2018simple}
Bin Xiao, Haiping Wu, and Yichen Wei.
\newblock Simple baselines for human pose estimation and tracking.
\newblock In {\em ECCV}, 2018.

\bibitem{yu2020humbi}
Zhixuan Yu, Jae~Shin Yoon, In~Kyu Lee, Prashanth Venkatesh, Jaesik Park, Jihun
  Yu, and Hyun~Soo Park.
\newblock {HUMBI}: A large multiview dataset of human body expressions.
\newblock In {\em CVPR}, 2020.

\bibitem{zhang2021pymaf}
Hongwen Zhang, Yating Tian, Xinchi Zhou, Wanli Ouyang, Yebin Liu, Limin Wang,
  and Zhenan Sun.
\newblock {PyMAF}: {3D} human pose and shape regression with pyramidal mesh
  alignment feedback loop.
\newblock In {\em ICCV}, 2021.

\end{thebibliography}
}

\end{document}